\documentclass[runningheads]{llncs}

\usepackage[T1]{fontenc}
\usepackage[utf8]{inputenc}
\usepackage{graphicx}
\usepackage{amsmath,amssymb}
\usepackage{url}
\usepackage{booktabs}
\usepackage{fancyhdr}

\title{Behavioral Universe Network (BUN): A Behavioral Information-Based Framework for Complex Systems}

\titlerunning{Behavioral Universe Network Framework Based on AIB}
\author{Wei Zhou, Ailiya Borjigin, Cong He}
\institute{Probe Group Pte. Ltd.\\
\{Zhou, Ailiya, Cong\_He\}@Probe-Group.com}

\begin{document}

\maketitle

\vspace{-1em}
\begin{center}
    \includegraphics[width=4cm]{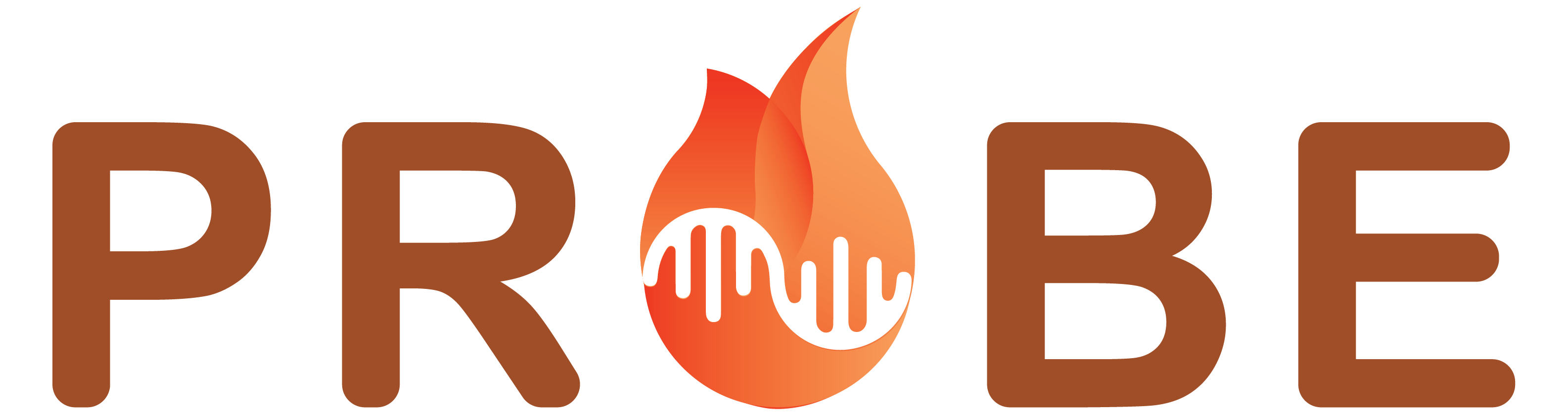}
\end{center}
\vspace{1em}

\begin{abstract}
Modern digital ecosystems are characterized by complex, dynamic interactions among autonomous entities across diverse domains. Traditional paradigms often treat agents and objects separately, failing to provide a unified theoretical foundation to capture their interactive behaviors. This paper introduces the \emph{Behavioral Universe Network} (BUN), a theoretical framework grounded in the \emph{Agent-Interaction-Behavior} (AIB) formalism. BUN treats \emph{subjects} (active agents), \emph{objects} (resources), and \emph{behaviors} (operations) as first-class citizens, all governed by a shared \emph{Behavioral Information Base} (BIB). We first detail the AIB core principles, defining how subjects, objects, and behaviors are formally described and regulated. We then describe BUN as a framework, showcasing how information-driven triggers, semantic object enrichment, and adaptive rules enable highly coordinated multi-agent systems. We highlight the framework's key advantages: more accurate behavior analysis, strong adaptability to dynamic environments, and cross-domain synergies. Finally, we outline open challenges and future work, positioning BUN as a promising foundation for next-generation digital governance and intelligent applications.
\end{abstract}

\section{Introduction}
\label{sec:intro}

Modern digital ecosystems are characterized by complex, dynamic interactions among autonomous entities across diverse domains. As governments and industries undergo rapid digital transformation, they face growing challenges in modeling and governing the behaviors of numerous interacting agents and systems \cite{SukhwalKankanhalli2022,Jennings2000}. Traditional paradigms of system design often treat agents and objects separately, without a unified theoretical basis to capture their interactive behaviors. This limits our ability to predict, coordinate, and adapt system behavior in complex scenarios \cite{Gaia2000}. Agent-based modeling (ABM) has emerged as a powerful technique to address some of these challenges by representing heterogeneous stakeholders and nonlinear interactions in digital governance ecosystems. However, a more fundamental framework is needed to formally integrate agents, their interactions, and resulting behaviors under one coherent model.

In this paper, we introduce the \textbf{\emph{Behavioral Universe Network} (BUN)}, a theoretical framework grounded in the \textbf{\emph{Agent-Interaction-Behavior} (AIB)} formalism, to address this need. BUN provides a principled way to represent \textbf{\emph{subjects} }(autonomous agents), \textbf{\emph{objects}} (resources or entities acted upon), and \textbf{\emph{behaviors} }(the interactions among subjects and objects) in a unified \textbf{\emph{Behavioral Information Base}}. By embedding semantic context into objects and capturing rich behavior traces and rules in an information base, BUN enables precise analysis and coordination of behaviors. This framework draws on and extends concepts from multi-agent systems, knowledge bases, and behavior modeling to create a deeply adaptive and cross-domain \textbf{behavioral network}. The AIB formalism, which underpins BUN, allows us to formally define and reason about the conditions under which agents (subjects) interact with objects to produce behaviors, ensuring that system rules and constraints are satisfied in every interaction.

The remainder of this paper is organized as follows. \textbf{Section~2} outlines the core principles of the AIB formalism, defining the subject--object--behavior triad and introducing the notion of a behavioral information base as a control mechanism. \textbf{Section~3} details the BUN framework driven by the behavioral information base, elaborating on the key components: (a) the Subject as an intelligent adaptive entity, (b) the Object as a semantically enriched behavioral object, and (c) Behavior as purposeful interactions mediated by information. \textbf{Section~4} describes the operational mechanisms of BUN, showing how coordination is achieved via the shared behavioral information base through (a) information-driven behavior triggers and propagation, and (b) the deep integration of models, data, and rules to realize a coordinated adaptive system. \textbf{Section~5} discusses the application advantages of BUN as a catalyst for multi-domain innovation: (a) enabling accurate behavior analysis and prediction, (b) providing strong adaptability and flexibility, and (c) fostering cross-domain synergies and expansion. Finally,\textbf{ Section~6} concludes the paper and outlines future research directions, positioning BUN in the context of ongoing developments in behavioral modeling, agent systems, and digital governance.

\section{Core Principles of the AIB Formalism}
\label{sec:AIBcore}
The \textbf{Agent-Interaction-Behavior (AIB)} formalism offers a conceptual and logical foundation for modeling digital interactions. Here, every \emph{behavior} is characterized by:
\[
   \text{Behavior} = \; S : f(O),
\]
where $S$ is the \emph{subject} (active agent), $O$ is the \emph{object} (resource acted upon), and $f$ is the operation or function performed. In other words, this logical underpinning ensures that any behavior in the system is permissible only if the agent, the object, and the intended action each meet their designated criteria. The formalism thereby provides a \textbf{policy envelope} for behaviors, ensuring consistency and compliance with system rules (such as security policies, business rules, or physical laws) whenever an interaction occurs.

\subsection{Subjects (Agents)}
A subject in AIB is an \emph{intelligent, autonomous entity} that can initiate behavior. It corresponds to actors such as human users, processes, software agents, or multi-agent organizations. In alignment with classical definitions of computing-system subjects~\cite{Bishop2003,Jennings2000}, a subject is ``an active entity capable of making requests and consuming resources.'' 
The AIB formalism extends this definition to incorporate \emph{intelligence and adaptability}~\cite{Wooldridge2009,Sutton2018}, meaning that a subject is expected to perceive context, choose behaviors that advance its goals, and potentially learn from outcomes to refine its decision processes.

\subsection{Objects (Resources)}
Objects are \emph{semantically enriched} information carriers or resources that can be manipulated by subjects. They remain passive in that they do not autonomously initiate actions~\cite{Jennings2000}. Examples include files, data records, devices, or environment resources. By embedding \emph{semantic descriptors} in a \emph{Behavioral Information Base}, each object can advertise its content, state, and relevant constraints, enabling more flexible agent-object interactions~\cite{Kifer1995}. As we will describe in Section~\ref{sec:BUNframework}, this semantic representation is key to achieving \emph{context-awareness} and \emph{cross-domain interoperability}.

\subsection{Behaviors (Operations)}
A behavior is a subject’s action on an object. Rather than viewing actions as implicit products of agents, AIB explicitly treats them as \emph{first-class entities}. This ensures that every behavior is \emph{accountable} and \emph{analyzable}. Following classical operating-system design~\cite{Lampson1974}, each behavior is associated with access-control checks, context validation, and outcome recording. A powerful property of the AIB approach is that it can unify policies, security, and logic checks into a single \emph{triple-based} structure:
\[
   \bigl(S \models P_1\bigr) \;\land\; \bigl(O \models P_2\bigr) \;\land\; \bigl(f(O) \models P_3\bigr) 
   \;\Longrightarrow\; \text{Behavior is valid}.
\]
Here $P_1, P_2, P_3$ are policy constraints on the subject, object, and the intended function, respectively.

\subsection{Behavioral Information Base}
A key concept in AIB is the \emph{Behavioral Information Base} (BIB), a repository storing:
\begin{itemize}
\item \textbf{Behavior histories}: Records of past interactions;
\item \textbf{Rules and policies}: Constraints $P_1, P_2, P_3$ that must be satisfied;
\item \textbf{Semantics and context}: Descriptors for objects and potential operations;
\item \textbf{Models of behavior}: Predictive or prescriptive models used by agents to plan interactions.
\end{itemize}
By consistently referencing the BIB, subjects \emph{discover feasible behaviors}, objects \emph{advertise constraints and affordances}, and the system ensures \emph{policy compliance}~\cite{Denning1982}. 

In summary, the AIB formalism establishes a \textbf{high-level logical model} of a digital environment: \textit{Subjects} (intelligent agents) engage in \textit{Behaviors} (interactions/operations) on \textit{Objects} (informational entities), subject to specific rules for each. This model is enriched and operationalized by the Behavioral Information Base, which contains the necessary knowledge to coordinate and constrain these interactions. By treating behavior as an explicit component and grounding it in an information base, AIB provides a bridge between \textbf{formal behavioral logic} and \textbf{practical system implementation}, forming the theoretical backbone of the Behavioral Universe Network framework.

\section{BUN Framework Driven by Behavioral Information Base}
\label{sec:BUNframework}

Building on the AIB formalism, the \emph{Behavioral Universe Network (BUN)} ~\cite{Zhou2025}framework instantiates these principles into an architectural and operational model for complex systems. The BUN framework is “driven by [a] Behavioral Information Base,” meaning that the generation, coordination, and adaptation of behaviors in the network are centrally influenced by information about behaviors. In BUN, every entity and action is \emph{behavior-centric}: subjects and objects are defined in relation to behavior, and the behavioral information base serves as the nexus linking them. We now describe the three fundamental components of BUN---\textit{Subject}, \textit{Object}, and \textit{Behavior}---as they are realized in this framework, highlighting their formal definitions (as per the digital asset model) and their enhanced roles in a BUN system.

\subsection{Subject: Intelligent Adaptive Entity}
In the BUN framework, a \textbf{Subject} is an intelligent, adaptive entity capable of purposeful action. This corresponds to the formal definition of subject in AIB as an active entity with autonomous behavior execution. In practical terms, a subject in BUN can be any actor such as a human user, a software agent, a cognitive robot, or even a composite organizational agent, provided it exhibits autonomy and goal-directed behavior. The subject is considered ``intelligent'' in the sense that it can perceive information, reason or compute decisions, and execute actions that pursue certain objectives~\cite{Panait2005}, and it is \emph{adaptive} in that it can modify its behavior based on experience or changing conditions---often achieved by learning from the Behavioral Information Base or updating its internal state. In AI terms, the subject aligns closely with the notion of an intelligent agent: ``an autonomous entity that perceives its environment, makes decisions, and takes actions to achieve specific goals''~\cite{Jennings2000}. Furthermore, an intelligent subject in BUN is expected to be capable of improvement---for example, using feedback to learn better actions---thus adapting over time~\cite{Schroeder2019}.

Each subject in BUN is associated with attributes that enable it to be an identifiable and accountable actor in the network. These include a unique identity or label, capabilities (what operations it can perform), and possibly roles or intent. According to formal definitions, the subject is the ``bearer of behavior''---behavior cannot exist without a subject to carry it out, and conversely an entity cannot be called a subject if it has no behavior~\cite{Zhou2025}. This reinforces that in BUN, being a subject implies \emph{agency}: the power to act and change the state of objects or the environment.

Crucially, a BUN subject is coupled to the Behavioral Information Base for guidance and coordination. The BIB contains the subject’s behavioral history (what it has done before), preferences or goals, and any rules it must obey (from $P_1$ in the AIB formalism). In essence, the BIB acts as the subject’s \emph{memory and knowledge repository}. For example, an autonomous vehicle agent (subject) in a BUN for smart traffic might consult the BIB for traffic rules (speed limits, right-of-way rules), learned patterns (historical data about congestion at certain times), and its own objectives (fastest route with safety constraints) before deciding how to behave at an intersection. The subject uses this information to select an appropriate behavior (e.g., slow down or stop) that fits the context. In a way, the Behavioral Information Base empowers the subject with \emph{informed autonomy}---the agent is autonomous, but its autonomy is informed by a rich base of knowledge and rules, ensuring that its actions remain within desired bounds.

Because subjects in BUN are intelligent and adaptive, they may also update the Behavioral Information Base as they operate. Each significant behavior or outcome can be recorded in the BIB as a new data point (a trace of behavior, result, success or failure, etc.), contributing to a growing \emph{behavioral knowledge repository}. This cumulative knowledge helps both the subject itself and other subjects: for instance, if one agent learns a more efficient way to perform a task, that information could be shared via the BIB for the benefit of others, continually improving the system’s performance. In summary, the \textbf{Subject} in BUN is the active decision-maker and actuator, characterized by intelligence, autonomy, and adaptability, and it is tightly integrated with the Behavioral Information Base that guides and constrains its actions.

\subsection{Object: Semantically Enriched Behavioral Object}
The \textbf{Object} in the BUN framework is defined as a \emph{semantically enriched behavioral object}. By definition, it retains the meaning of an object from AIB: a passive entity that is acted upon by subjects and which carries or embodies information~\cite{Zhou2025}. However, BUN extends this concept by endowing objects with \emph{semantic descriptors} and \emph{behavioral interfaces} that make them active participants in the behavioral network \emph{in an informational sense}, even if they are not autonomous agents. In other words, while an object does not initiate behavior on its own, it provides rich information that influences how behaviors can be performed on it. Each object in BUN is described in the Behavioral Information Base with metadata capturing its semantics (what the object represents, its type or class), its state (current condition or data content), and the affordances or operations that can be applied to it.

This semantic enrichment draws from the ideas of the Semantic Web and IoT, where objects (or ``things'') are annotated with machine-understandable information to enable smarter interactions. For example, consider a document in a digital governance system (the object). In a traditional system, the document might just be a file with an ID. In BUN, the document object could be annotated in the BIB as, say, ``Type: FinancialReport; Sensitivity: Confidential; Status: Approved; LinkedPolicy: Must not be released publicly before date X.'' Such semantic information means that any subject wanting to act on this object (e.g., to share it or modify it) can query the BIB and immediately know the context and constraints. The object’s semantics thus directly inform subject behavior. A subject agent attempting a behavior not permissible by the object’s metadata (like releasing the report early) would find a rule violation in the BIB and could be prevented or discouraged from that action.

Formally, each object must satisfy the object-specific rules $P_2$ for any behavior to occur~\cite{Zhou2025}. These rules are often derived from the object’s attributes and semantics. The BUN framework centralizes these rules in the Behavioral Information Base. In practice, this could be implemented as ontologies or schemas that describe object categories and their allowed interactions. The notion of ``behavioral object'' also implies that objects can have associated \emph{behavioral models}. While objects themselves do not act, the system may have models predicting how the object might change under certain actions (for instance, a data object might have a validation model to check integrity after an update, or a physical object might have a model of wear-and-tear if used by agents in a simulation). These models are stored in the BIB as well, thus a subject planning an interaction can consult not just static rules but also \emph{model-based predictions} of outcomes.

Enriching objects with semantics and models confers several advantages. It greatly improves \emph{interoperability} and integration of heterogeneous components, as the shared semantic schema allows different agents (potentially from different organizations or domains) to understand objects in a common way~\cite{Kolozali2019}. It also paves the way for \emph{automation}---agents can reason about objects they have never seen before if those objects are described in standard semantic terms. This property is vital in cross-domain systems where, for example, an IoT sensor (object) might be accessed by a cloud service agent (subject); a semantic description of the sensor’s data format and meaning allows the agent to use it appropriately without custom programming. BUN’s use of semantically enriched objects is akin to giving objects a ``voice in their own handling''—the object, through its metadata in the BIB, ``tells'' the agent what it is and how it can be used.

In summary, the \textbf{Object} in BUN remains a passive information carrier and target of actions, but it is \emph{behaviorally aware} in that its properties and constraints are explicitly recorded and utilized in the Behavioral Information Base. This ensures that interactions involving the object are context-aware and semantically correct. By formally integrating object semantics into the framework, BUN enables a higher degree of automation, correctness, and cross-domain compatibility in agent-object interactions than traditional systems.

\subsection{Behavior: Purposeful Interactions Based on Information Base}
In the BUN framework, \textbf{Behavior} is elevated to a first-class component: it represents the \emph{purposeful interactions} among subjects and objects, orchestrated by and recorded in the Behavioral Information Base. Every behavior in BUN is not just a blind action; it is generated \emph{based on information}—specifically, information from the BIB about goals, context, and rules. This ensures that behaviors are \emph{deliberate} (goal-oriented) and \emph{constrained} (rule-compliant).

A behavior in BUN typically follows a cycle: a subject has a certain goal or is triggered by an event, it queries the Behavioral Information Base to gather relevant information (about the object, environment, and any governing rules or past similar behaviors), then it executes an interaction with the object in line with that information, and finally the outcome of the behavior is fed back into the information base (closing the loop with learning or trace logging). Through this cycle, \emph{information drives behavior, and behavior updates information}, creating a continuous feedback mechanism that underpins system adaptability.

The term ``purposeful'' underlines that behaviors are linked to objectives or functions. In cognitive agent terms, a BUN behavior often aligns with an agent’s plan to achieve a specific goal state. We can draw a parallel to the BDI (Belief-Desire-Intention) model of agency, where an agent’s desires (goals) and beliefs (information base) produce intentions (committed plans) which result in actions. BUN can be seen as providing the ``belief'' part (a shared information base) and a formal envelope so that the intentions (behaviors) are carried out in a consistent way. A behavior might be simple (a single action like ``Agent A reads Data B'') or composite/complex (a protocol or transaction consisting of multiple steps and interactions among multiple agents). In all cases, BUN attempts to capture the semantics of the behavior in the BIB—for instance, labeling a composite behavior with its purpose (``data backup operation'') and recording its steps.

One important aspect of BUN behaviors is that they are \emph{information-driven}. This means external stimuli or internal states encoded in the information base serve as triggers or preconditions for behaviors (elaborated further in Section~4). For example, a rule in the BIB might state: ``If sensor reading X exceeds threshold Y, then behavior Z (\texttt{emergency shutdown}) should be performed.'' The availability of this rule in the information base makes the behavior \emph{responsive} to environmental information. Because the information base is shared or accessible across the network, multiple agents can coordinate around these triggers.

Behaviors in BUN are also closely monitored and regulated via the information base. Recall from the AIB formalism that a behavior $a = S:f(O)$ must satisfy the compound rule $P_1 \land P_2 \land P_3$ to be considered valid~\cite{Zhou2025}. In BUN, the enforcement of this is done by consulting the BIB at runtime. Before or during a behavior, the system can check: does the subject satisfy $P_1$? (e.g., does this user have the role required to perform this operation), does the object satisfy $P_2$? (e.g., is the object in a state that allows writing), and are the contextual conditions $P_3$ met? (e.g., is this operation allowed at this time or under current system load?). Only if all are true does the behavior proceed, so the behavior is inherently \emph{policy-compliant} by design. This approach is more powerful than hard-coding checks in each application because the rules are centralized in the BIB and can be updated or expanded without changing the agents’ code.

Finally, purposeful behavior also implies \emph{measurable outcomes} and potential for \emph{prediction}. By logging behaviors and their outcomes in the BIB, the BUN framework can accumulate data to analyze how behaviors influence system state over time. This can be used to refine future behaviors (making them more effective toward goals) or to predict likely future actions (by subjects) and effects (on objects) under certain conditions. For instance, if a smart grid system (implemented as BUN) records that ``behavior: reduce load on transformer when temperature $>$ 80$^\circ$C'' always prevents overheating, it can predict that in future high-temperature scenarios, a load-shedding behavior will occur and the transformer will stay safe. Thus, the purposeful behaviors in BUN not only achieve immediate goals but also contribute to a \emph{self-refining loop of behavioral knowledge}, improving the system’s ability to plan and coordinate complex tasks.

In summary, \textbf{Behavior} in BUN is treated as a \emph{goal-driven, information-informed interaction}. Every behavior is explicitly linked to information (conditions, triggers, and rules in the BIB) and is executed to fulfill a defined purpose. By grounding behaviors in the Behavioral Information Base, the BUN framework ensures that what agents do is neither random nor rigidly pre-programmed, but rather dynamically determined by the current context and accumulated knowledge. This paves the way for highly adaptive and reliable system behavior, even as conditions evolve.

\section{Operational Mechanisms of Behavioral Information Base (BIB)}
\label{sec:Operational}

A key strength of the Behavioral Universe Network framework lies in its operational mechanisms that enable \emph{system-wide coordination} of behaviors. Unlike isolated agent systems, BUN assumes that agents (subjects) are working in a \emph{networked environment} where their actions may affect one another and must often be harmonized toward broader system objectives. The \emph{Behavioral Information Base} (BIB) is the cornerstone for this coordination: it serves as the medium through which information is shared and through which behaviors are orchestrated. In this section, we examine two critical aspects of BUN’s operation: 
\begin{enumerate}
    \item How behaviors are \emph{triggered and propagated} in an \emph{information-driven} manner;
    \item How \emph{models, data, and rules} in the information base facilitate \emph{deep coordination}, making the system adapt in a coherent way rather than as disparate parts.
\end{enumerate}

\subsection{Behavior Trigger and Propagation: Information-Driven Dynamics}
In BUN, the initiation of behaviors across the network is largely \emph{event-driven} and \emph{information-driven}. This means that changes in the information base (or incoming data from the environment) act as triggers that cause one or more agents to execute behaviors. The paradigm is analogous to a blackboard system in AI, wherein multiple specialists (agents) collaboratively solve a problem by reacting to information posted on a common blackboard~\cite{Nii1986,Erman1980}. Here, the Behavioral Information Base acts as a \emph{blackboard}: when a subject writes a piece of information (e.g., an alert, an update, a request) or a change in object state is recorded, other subjects monitoring the BIB may detect conditions that prompt them to act~\cite{Freedman1982}. This leads to a \emph{propagation of behaviors}---one agent’s action leads to changes that induce another agent’s action, and so on, potentially forming a cascade or a carefully choreographed sequence.

\smallskip
\noindent
\textbf{Example:} In a smart city context, a traffic sensor (object) detects an accident on a highway and updates a shared incident log in the BIB. This update serves as a trigger; nearby autonomous vehicles (subjects) query the BIB regularly for such incidents and, upon finding the new entry, each initiates a behavior to slow down or reroute. Simultaneously, a city traffic control agent sees the same incident in the BIB and activates a behavior to change digital road signs and notify emergency services. In this scenario, \emph{one piece of information} (the accident event) triggered multiple coordinated behaviors across different agents, all mediated through the shared information base. This exemplifies \emph{information-driven dynamics}---the flow of information dictates the flow of behavior, rather than behaviors happening in isolation.

To manage such dynamics, BUN employs a \emph{publish/subscribe} or \emph{notification} mechanism on top of the BIB. Agents can subscribe to certain information patterns or event types in the BIB (e.g., ``subscribe me to any change in object X’s state'' or ``notify me when rule Y’s conditions are satisfied''). When the relevant information appears or changes in the BIB, the subscribed agents are alerted and can decide to act. This is effectively how behavior triggers are implemented. It is akin to how event-driven architectures in software operate, where components react to events. However, in BUN, the \emph{events and states are recorded in a semantically rich information base} that all agents reference. Research in multi-agent systems emphasizes that beyond simple message passing, \emph{structured interaction protocols and shared context} are needed for effective coordination~\cite{Corkill1979}. BUN’s approach of using a shared information hub aligns with these findings, providing a structured yet flexible way for agents to coordinate via data.

Propagation of behavior in BUN is carefully controlled to avoid chaotic cascades. The BIB can include meta-rules to throttle or direct propagation. For example, it might contain a rule: ``If event A triggers behavior B in agent X, suppress duplicate triggers to avoid oscillation.'' This is important in closed-loop systems to maintain stability. The information base can thus act not only as a \emph{trigger source} but also as a \emph{governor}, ensuring that triggers lead to productive behavior chains that converge toward a solution or steady state~\cite{Freedman1982}. In the blackboard analogy, this is like the \emph{control shell} that moderates which agent should act next based on the blackboard state~\cite{Nii1986}; in BUN, the control logic can be embedded as rules in the BIB that coordinate sequence and priority of behaviors.

Moreover, behavior propagation in BUN often involves \emph{multicast} of information: one event is relevant to many agents. The semantic structure of information in the BIB enables agents to filter what is relevant. For instance, an update labeled with a geographic tag will only trigger agents in that region. This selective propagation prevents overwhelming unrelated parts of the network with unnecessary triggers. Essentially, contextual information stored with events (who, where, when, what) allows \emph{targeted coordination}.

In summary, \emph{Behavior Trigger and Propagation} in BUN is an \emph{information-centric process}. The Behavioral Information Base plays an active role: it not only houses static rules but also dynamically evolving state that serves as the stimulus for agent behaviors. Agents continuously sense this information environment (like observing a blackboard) and respond when conditions match their trigger criteria. This leads to a responsive, event-driven system where \emph{coordination emerges from shared situational awareness} provided by the BIB. The result is that BUN can handle complex, distributed scenarios (such as emergency response, automated trading systems, or collaborative robotics) with a high degree of synchronization and coherence among the behaviors of different agents, all driven by the timely flow of information.

\subsection{Models, Data, and Rules: Deeply Coordinated Adaptive System}
Coordination in BUN is not only about reacting to immediate events; it is also about the \emph{deep integration} of knowledge (models), real-time data, and governing rules to achieve a \emph{harmonious adaptive system}. The Behavioral Information Base is the repository of these three critical elements: it contains \emph{data} (current states, events, history), \emph{models} (predictive or normative models of behavior, possibly learned or engineered), and \emph{rules} (constraints, policies, logic). The synergy of these elements in one place is what allows BUN to coordinate behaviors at a profound level, enabling complex adaptation that is \emph{system-wide} rather than just local to each agent.

\paragraph{Models.}
Models in the BIB refer to any formal or learned representations that can predict or prescribe behavior. These could be machine learning models, simulations, or mathematical formulations of processes. For example, a model could predict how a crowd will behave in an evacuation, or how an IT system’s performance will change under certain loads. By storing such models in the BIB, BUN ensures all agents can \emph{share a common understanding} of dynamic processes. Coordination is enhanced because each agent is not operating on its own guesswork; they draw from a consistent model. For instance, if multiple agents are managing different subsystems of a smart grid, a load distribution model in the BIB can help them coordinate electricity usage without conflict, as they all refer to the same model outputs when deciding their actions. This addresses the challenge identified in multi-agent research that \emph{common knowledge can greatly improve coordinated decision-making}. When agents know certain facts or models are commonly known, they can plan with the expectation that others will act in accord, achieving more complex decentralized coordination.

\paragraph{Data.}
\emph{Data} in the BIB ensures that coordination is grounded in reality. Real-time data feeds (from sensors, logs, user inputs) update the BIB and thereby inform agent behaviors. Beyond raw data, the BIB aggregates and contextualizes data---possibly through streaming analytics or data fusion. Because the data from various sources resides together, the system can detect \emph{cross-cutting patterns} that individual agents might miss. For example, data from different parts of a supply chain (inventory levels, shipping delays, demand surges) can all be in the BIB; a coordination mechanism (perhaps a model or rule) can then adjust the behaviors of procurement agents, logistics agents, and sales agents in concert to mitigate a supply shock. The shared data context prevents agents from working at cross purposes with outdated or siloed information. It creates a \emph{common operational picture} in organizational contexts, which is crucial for coordinated response.

\paragraph{Rules.}
\emph{Rules} in the BIB, including both fixed policies and dynamically created rules, function as the \emph{grammar of coordination}. They ensure not only that individual behaviors obey constraints, but also that multi-agent interactions follow agreed protocols~\cite{Smith1980}. For example, a rule might enforce that two agents cannot write to the same object simultaneously (to avoid inconsistency), thereby coordinating via mutual exclusion. Another rule set might define a negotiation protocol: ``Agent A must respond to Agent B’s request within 5 seconds or yield the task to Agent C.'' By encoding these interaction protocols centrally, BUN avoids the pitfalls of purely emergent systems where agents might otherwise have to develop their own coordination via trial-and-error. Instead, coordination mechanisms are \emph{designed into the fabric} of the system via the BIB rules. In BUN, these protocols are effectively implemented as rule-based workflows in the BIB that agents follow.

\medskip
The \emph{deep coordination} enabled by BUN arises when \emph{models, data, and rules} work in unison. Suppose a sudden change happens in the environment; data enters the BIB reflecting that change. A model in the BIB processes this data to forecast implications, and then rules in the BIB trigger a set of behaviors across agents to adapt to the forecast. This entire loop can happen quickly and repeatedly, yielding an \emph{adaptive system} that is both \emph{proactive} and \emph{reactive}. It is \emph{reactive} to the immediate data (current events) and \emph{proactive} via the model (anticipating future states), with rules ensuring coherent execution. This kind of tight integration is what we mean by ``deeply coordinated'': coordination is not an afterthought or just emergent from agent interactions, but rather an \emph{ingrained capability} of the system designed through the BIB content.

An illustrative example can be drawn from \emph{autonomic computing} (self-managing systems). In an autonomic resource management system (like cloud infrastructure), BUN could be applied such that the BIB holds rules for maintaining performance (SLA policies), data from monitoring (CPU, memory usage, response times), and models for workload forecasting. When usage spikes (data), the model predicts a potential overload, and a rule triggers scaling behaviors (various agents launching new server instances, load balancers reconfiguring) in a coordinated fashion to preempt any failure. The result is a flexible yet controlled adaptation to load changes, all mediated by shared knowledge. Traditional systems might have independent modules making decisions that conflict (e.g., two managers trying to scale up redundantly or not at all), whereas BUN’s integrated approach fosters a \emph{unified response}.

In summary, through the \emph{Behavioral Information Base}, BUN achieves a level of coordination where the \emph{collective behavior} of the system is more than the sum of individual agent behaviors. Models provide foresight and consistency, data provides situational awareness, and rules provide structured interaction patterns. Together, these enable a highly adaptive system where agents act in concert, guided by a common informational and normative substrate. This approach greatly reduces misalignment and inefficiencies that can plague complex multi-agent or distributed systems, making BUN-operated systems more resilient, efficient, and predictable even in the face of changing conditions or requirements.

\section{Application Advantages: Enabling Multi-Domain Innovation}
\label{sec:Applications}

The Behavioral Universe Network (BUN) framework offers several compelling advantages that act as a catalyst for innovation across multiple domains. By rethinking system architecture around \textit{behavior and information}, BUN addresses longstanding challenges in analyzing complex behaviors, building adaptable systems, and achieving interoperability between domains. In this section, we highlight three major advantages of the BUN approach: 
\begin{enumerate}
    \item Accurate behavior analysis and prediction,
    \item Strong adaptability and flexibility of systems, and
    \item Enhanced cross-domain synergies and expansion opportunities.
\end{enumerate}

These benefits position BUN not only as a theoretical construct, but as a practical enabler for advanced applications in areas such as smart governance, autonomous systems, industrial internet, and beyond.

\subsection{(a) Accurate Behavior Analysis and Prediction}

Understanding and predicting behaviors in complex systems—whether human, artificial, or hybrid—has always been difficult. Traditional data analytics might capture statistics, and simple simulations might handle small agent sets, but as interactions grow in scale and complexity, analysis becomes less effective. BUN provides a structured way to log and analyze behaviors via the Behavioral Information Base (BIB). Every interaction is recorded with context, turning the system into a \textit{behavior observatory}, where every action is observable and semantically annotated.

This structured behavior capture directly enables predictive analytics. Techniques from behavior modeling and social simulation can be applied to BUN data to discover patterns and build predictive models~\cite{Schroeder2019}. For instance, one might apply sequence mining to identify common failure chains, or train a machine learning model to predict an agent’s next action based on state and past behavior~\cite{Kolozali2019}. Because the BIB also stores policies and constraints, predictions generated on top of it can better comply with real-world operational rules.

In digital governance, this capability is essential. Government platforms that interact with both citizens and AI agents benefit from BUN's ability to model citizen-service interactions and forecast demand patterns~\cite{SukhwalKankanhalli2022}. The BUN formalism can calibrate and enhance traditional ABM-based models used for policy simulation and service optimization. 

In cybersecurity, BUN supports anomaly detection and proactive security. By modeling user and system behaviors in BIB, deviations from normal behavioral sequences can be detected early. This enhances behavioral analytics used in threat detection~\cite{Laskov2011}. As a comprehensive behavior log, the BIB enables correlation of events across subsystems, improving the accuracy of intrusion detection and insider threat identification.

In summary, BUN enables high-resolution behavioral data capture and structured interpretation, which significantly improves the ability to analyze past events and forecast future behavior. This helps stakeholders move from reactive to \emph{predictive and preventive management}.

\subsection{(b) Strong Adaptability and Flexibility}

Adaptability—the capacity to handle change—is critical in modern digital environments. BUN inherently supports adaptability due to its information-driven behavior generation and decoupling of logic and policy.

First, behavior is not hard-coded. If a new type of event or object appears, agents can query the BIB for instructions or matching patterns. Once rules are updated, all agents operate under the new regime without being individually reprogrammed. This \emph{policy-based adaptation} allows fast system-wide changes.

Second, BUN systems can \emph{learn and self-tune}. As agents execute behaviors and outcomes are stored, the BIB becomes a growing behavioral memory. Reinforcement-like learning can emerge: agents favor behaviors with successful histories, which improves overall system performance. Since the knowledge is shared, one agent’s learning benefits all~\cite{Abowd2000}.

Third, flexibility in BUN refers to modularity. Agents, objects, and rules are loosely coupled. New agents can join and operate as long as they conform to the BIB. This is particularly important in \textit{evolving ecosystems} such as smart cities, where system components are continuously upgraded.

BUN also supports graceful degradation and recovery. If a component fails, others can assume its role using BIB state. For example, if Agent A fails, Agent B can inspect the BIB, see that A's task is incomplete, and continue it if permitted. This enables dynamic task reallocation without central orchestration.

In summary, BUN provides \emph{live adaptivity} and \emph{runtime flexibility}, reducing lifecycle costs, improving robustness, and outperforming traditional static architectures when facing environmental volatility~\cite{Sommer2010}.

\subsection{(c) Cross-Domain Synergies and Expansion}

A powerful benefit of BUN is its ability to \textit{bridge multiple domains}. Many of today’s critical problems span sectors—e.g., energy and transportation, health and finance. Yet integration is difficult due to incompatible data formats, vocabulary, and control policies.

BUN overcomes this through a \emph{universal behavior formalism} (AIB) and a semantic BIB shared across agents. It provides a \emph{common language of behavior}—a digital lingua franca that domains can use to describe and exchange behavior logic~\cite{Bandini2008}.

\textbf{Example:} In a smart city, energy and transport are typically managed separately. But by using BUN, both can be modeled in AIB: agents represent grid controllers and vehicles, objects represent power or road segments, and behaviors represent charging, discharging, routing, etc. Shared BIB rules, such as ``allow EVs to feed energy back when frequency drops,'' allow synergy that would be impossible under siloed architectures.

Another example is digital governance interacting with fintech or social platforms. With proper privacy safeguards, behavioral signals from citizen financial activities could trigger support policies. This matches calls in digital governance research for better ABM + knowledge base integration~\cite{SukhwalKankanhalli2022,Park2020,Gama2013}.

\textbf{Scalability.} The BUN framework is abstract and recursive. You can start with one domain and gradually integrate others by bridging their BIBs or translating their ontologies. Eventually, this forms an \textit{ecosystem of BUNs} that cover entire segments of digital society. Semantic standardization of BIB rules makes this viable.

The synergy effect means \emph{emergent services} can arise across domains. For example, a ``digital twin'' composed of health, finance, and social domains could produce holistic well-being insights for users. This is possible only when those domains interoperate at the behavioral level, as enabled by BUN.

\textbf{Summary.} BUN’s unifying abstraction—subject, object, behavior—and its shared information substrate (BIB) offer a foundation for composable, interoperable systems. It is well-suited to modern needs for cross-domain coordination and continuous integration.

\section{Conclusion and Future Work}
\label{sec:Conclusion}
This paper introduced the \textbf{Behavioral Universe Network (BUN)} as a \emph{behavior-centric framework} built upon the \textbf{Agent-Interaction-Behavior (AIB)} formalism. We showed how \emph{subjects}, \emph{objects}, and \emph{behaviors} are organized within a unified \emph{Behavioral Information Base}, enabling policy-enforced interactions and deep coordination. By modeling the system around behaviors and applying an information-driven approach, BUN offers:
\begin{itemize}
\item A comprehensive foundation for multi-agent action logging, rule checking, and semantic integration;
\item Strong adaptability, facilitated by centralized knowledge updates and machine learning synergy;
\item Opportunities for cross-domain interoperability, supporting complex scenarios that require heterogeneous agents to collaborate.
\end{itemize}

\noindent
\textbf{Future Work.}
Implementing real-world BUN systems requires further investigation into:
\begin{itemize}
\item \emph{Scalable BIB infrastructure}: Distributed data management, especially for large-scale IoT or cross-enterprise settings;
\item \emph{Domain-specific ontology development}: Ensuring consistent semantics for objects and behaviors across different application areas;
\item \emph{Advanced learning and optimization}: Integrating online machine learning and reinforcement learning for continuous behavior adaptation;
\item \emph{Governance and ethics}: Providing transparent human oversight mechanisms to manage BUN-based coordination responsibly~\cite{Sukhwal2022}.
\end{itemize}
Given the rising complexity of digital ecosystems, we believe BUN represents a powerful step toward robust, adaptive, and integrated \emph{behavioral governance} in next-generation intelligent systems.


\section*{Acknowledgments}

We dedicate this paper to the memory of Professor Yanwen Qu , whose seminal work, \emph{The Reappearance of Life in the Network World: Traditional Chinese Medicine and Modern Biological Behavior and Systems Science }, provided foundational inspiration and theoretical underpinnings for this research.

Professor Qu served as the Executive Vice Chairman of the Information Security Industry Association, China Information Industry Chamber of Commerce, an expert advisor to the China Information Security Evaluation Center, and as an adjunct professor at prestigious institutions including Peking University, Wuhan University, and Huazhong University of Science and Technology. He was also an esteemed member of China's Computer Science Group and a pioneering figure in the fields of computer-aided design (CAD), software engineering, computer science, and information security in China.

Although Professor Qu has passed away, his profound contributions, visionary insights, and mentorship continue to inspire and deeply influence our academic pursuits and research directions.

\end{document}